\documentclass{article}

\usepackage[hmarginratio=1:1,top=32mm]{geometry} 
\usepackage{amsmath}
\usepackage{graphicx} 
\usepackage[table,dvipsnames]{xcolor}
\usepackage{natbib}
\usepackage[colorlinks=true, citecolor=brown, linkcolor=brown, urlcolor=brown]{hyperref}
\newcommand{\hlink}[2]{\href{#1}{\underline{\smash{#2}}}} 

\usepackage{tikz}

\usepackage{gb4e}
\noautomath
\title{\textbf{Why Linguistics Will Thrive in the 21st Century:\\A Reply to Piantadosi (2023)}} 
\author{\textbf{Jordan Kodner}, \textbf{Sarah Payne}, \textbf{Jeffrey Heinz}\\
    Department of Linguistics, and\\
  Institute for Advanced Computational Science (IACS) \\
    Stony Brook University\\  
  {\tt \{first.last\}@stonybrook.edu}\\
}
\date{}

\begin{document}

\maketitle

\section{Introduction}

We present a critical assessment of \citeauthor{piantadosi2023modern}'s (\citeyear{piantadosi2023modern})
claim that ``Modern language models refute Chomsky's approach to language,'' focusing on four main points.
First, despite the impressive performance and utility of large language models (LLMs),
humans achieve their capacity for language
after exposure to several orders of magnitude less data. 
The fact that young children become competent, fluent speakers of their native languages
with relatively little exposure to them
is the central mystery of language learning
to which Chomsky initially drew attention, and LLMs currently show little promise of solving this mystery. 
Second, what can the artificial reveal about the natural?
Put simply, the implications of LLMs for our understanding
of the cognitive structures and mechanisms underlying language and its acquisition
are like the implications of airplanes for understanding how birds fly. Third, LLMs cannot constitute scientific theories of language for several reasons, not least of which is that scientific theories must provide interpretable explanations,
not just predictions. This leads to our final point: to even determine whether the linguistic and cognitive capabilities of LLMs rival those of humans requires explicating what humans' capacities actually are. In other words, it requires a \textit{separate} theory of language and cognition; generative linguistics provides precisely such a theory.
As such, we conclude that generative linguistics as a scientific discipline will remain indispensable throughout the 21st century and beyond.

\section{``Unconstrained'' Learning from Big Data is Not Human}
\label{small-data}

OpenAI's newest product at the time of writing, GPT-4,\footnote{\hlink{https://web.archive.org/web/20230314174836/https://openai.com/research/gpt-4}{https://web.archive.org/web/20230314174836/https://openai.com/research/gpt-4}. GPT stands for ``Generative Pre-trained Transformer.''}
performs very well on a wide range of standardized tests (though see \citealt{martinez2023re}). 
For example, on the LSAT, OpenAI reports that GPT-4 scored
better than roughly 88\% of human test takers
seeking admission to American law schools,
a significant boost from last year's OpenAI product, GPT-3.5,
which only outperformed about 40\% of human test takers.
Both the performance and pace of improvement are remarkable achievements. 
Nonetheless, the fact remains that young adults who pass the LSAT
are doing so without having read the trillions of sentences
and structured internet data that
these large language models are trained on. 
The difference in training regimes is stark
and highlights the fundamental question:
how do humans come to pass the LSAT, and other standardized tests,
on comparatively so \textit{little} data?

Parallel questions arise in the domain of language acquisition by children, which is characterized by a relative paucity of linguistic experience.
Children are exposed to at most about ten million tokens per year \citep{hart1992american,gilkerson2017mapping},
and most children have vocabularies of under one thousand words around age three,
regardless of the language being learned
\citep{bornstein2004cross, fenson1994variability}.
Yet these same children produce sentences
that largely obey the grammatical rules of their communities' languages
\citep{berko1958child, brown1973first, montrul2004acquisition, yang2006infinite, phillips2010syntax, Slobin2022}.
Thus, a fundamental question of linguistic study is
how children become fluent in their native language(s)
at a young age from so little data and experience. 

The disconnect between the linguistic experience (input)
and the linguistic capacity (output) is what gives rise to
the \textit{The Poverty of the Stimulus} argument
for the hypothesis that many aspects of language learning and representation are innate
\citep{chomsky1959review,chomsky1980cognitive,NowakEtAl2001,Yang2013}.
Under this hypothesis, children generalize from their limited input
in specific ways, navigating a constrained space
of possible natural language grammars. 
Consequently, they do not consider all logically possible generalizations
that are consistent with their linguistic experience.  
Rather, the particular structure of
the hypothesis space
facilitates the rapid development of their linguistic capabilities. The Poverty of the Stimulus is not tied to a specific theory of language, such as Minimalism or particular variants thereof, but rather follows from the basic problem of making generalizations from experience, as we describe in the next section.

The contrast between the input to the child and the input to LLMs is striking,
but Piantadosi is not concerned by this discrepancy.
He makes two claims.
The first is that LLMs refute so-called nativist theories of language because
``modern language models succeed despite the fact
that their underlying architecture for learning is relatively unconstrained''
\citet[18]{piantadosi2023modern}. 
In other words, he argues that broadly ``blank slate'' approaches to language learning
are in fact viable, with LLMs serving as a proof of concept. 
The second claim is  that ``our methods for training [LLMs] on very small datasets will inevitably improve'' \citep[14]{piantadosi2023modern}. From these two claims, Piantadosi concludes that LLMs show that unconstrained learning from small data is possible. The remainder of this section addresses each argument in turn.

\subsection{Feasible Learning Must be Constrained}


Piantadosi is not the first researcher to claim that the general architecture of LLMs, or deep artificial neural networks (ANNs) more broadly, ``is relatively unconstrained.'' As \citet[5]{baroni2022proper} points out,
 work situating deep networks ``within a broader theoretical context [does so] invariably in terms of nature-or-nurture arguments resting on a view of deep nets as blank slates." 
For example, \citet[637]{warstadt2019neural} take the position that that ``if linguistically uninformed neural network models achieve human-level performance on specific phenomena...this would be clear evidence limiting the scope of phenomena for which the [argument of the Poverty of the Stimulus] can hold." \citet[43]{pater2019} similarly claims that 
``with the development of the rich theories of learning represented by modern neural networks, the learnability argument for a rich [Universal Grammar] is particularly threatened."
We refer the reader to \citet{baroni2022proper} for a plethora of further examples of such claims drawn from the literature.

If deep ANNs really were so unconstrained, however, why would machine learning scientists constantly tinker with the layers, the gating mechanisms, the architectures, and the tuning of the hyperparameters? The reality is that these systems are biased in ways that are not well-understood. Paraphrasing a turn of phrase from \citeauthor{rawskiheinz2019}'s (\citeyear{rawskiheinz2019}) critique of \citet{pater2019}, ``Ignorance of bias does not imply absence of bias.'' 
Indeed, \citet{kharitonov2020they} found that when deep ANNs were trained on a small dataset which could have been generated by either a hierarchical or linear function, LSTMs with attention and Transformers apparently inferred a hierarchical function, while LSTMs without attention and CNNs inferred a linear one. Such a result indicates the presence of robust biases in the network architecture and certainly does not support a ``blank-slate" view of deep networks.
Though Piantadosi cites \citet{baroni2022proper} in support of his claim that LLMs constitute linguistic theories (\S \ref{not-theory}), he fails to note the second half of Baroni's claim: that deep networks ``are linguistic theories, \textit{not} blank slates" (pg. 6, emphasis ours). As \citet[7]{baroni2022proper} argues, 
 ``it is more appropriate, instead, to look at deep nets as\ldots encoding non-trivial structural priors facilitating language acquisition and processing.'' 
 Such ``non-trivial'' priors are fundamentally at odds with 
 a view of LLMs as ``relatively unconstrained.'' 

But \textit{even with such non-trivial priors}, 
the success of current LLMs depends at least in part 
on being trained on inhumanly large amounts of data \citep[e.g.,][]{kaplan2020scaling}. 
Indeed, results from the field of computational learning theory (CLT) have established that the kind of ``relatively unconstrained'' learning Piantadosi suggests is not possible given feasible computational resources, in terms of time and data.  
CLT studies
what it means to learn a concept from experience from a formal mathematical perspective.
The primary conclusion from this research is that
there are fundamental computational laws of learning that cannot be shortcut. 
Informally, these laws say that there is a trade-off between
the \textit{family of concepts} that one wishes to learn,
the \textit{kinds of data} one wishes to learn those concepts from,
and the \textit{computational resources} (time and space)
within which one needs to accomplish that learning. 
In particular, it is not possible to learn all computable concepts
from arbitrary data presentations representative of them,
with feasible amounts of computational resources.
This central result is encountered again and again
as researchers study different definitions of what ``learning'' means and
examine which families of concepts are learnable under such definitions
\citep{gold67,WolpertMacready1997,vapnik::statistical::1998,JainEtAl1999,niyogi2006,DeRaedt2008,MRT2012,Valiant2013}.

\begin{figure}[t!] 
\centering
\begin{tikzpicture}
  \begin{scope}[blend group = screen]
    \fill[black!50!white]  (0,6) rectangle (10,10);
    \fill[Turquoise!70!white] (0,6) rectangle (2.5,0);
    \fill[Purple!70!white]   (5,0) rectangle (7.5,6);
    \fill[Goldenrod!80!white]   (2.5,0) rectangle (5,6);
  \end{scope}
  \draw[black, thick] (0,0) rectangle (10,10);
  \draw[black, thick] (0,6) rectangle (10,0);

  \draw[black, thick] (2.5,0) rectangle (5,6);
  \draw[black, thick] (2.5,0) rectangle (7.5,6);
  \filldraw[black] (3.5,3) circle (1pt) node[anchor=west]{$(d, c)$};

  \draw (-1.5,5) node[rotate=90] {Logically Possible Concepts};
  \draw (-0.5,8) node[rotate=90] {Uncomputable Concepts};
  \draw (-0.5,3) node[rotate=90] {Computable Concepts};

  \draw (2.5,-0.5) node {$\underbrace{\hspace{5cm}}_{\text{\normalsize Positive Data Presentations}}$};
  \draw (5,-1.5) node {$\underbrace{\hspace{5cm}}_{\text{\normalsize Computable Data Presentations}}$};
  \draw (5,-2.5) node {Logically Possible Data Presentations};
\end{tikzpicture}

\caption{A visualization of how learning problems are defined. Defining a learning problem requires defining which kinds of concepts ($y$-axis) ought to be obtained from which kinds of data ($x$-axis). Another important parameter (not shown here) are computational complexity restrictions on the learning mechanism itself. The yellow rectangle exemplifies some choices in formulating a learning problem. Point $(d,c)$ is an instance of the problem: learning concept $c$ from data presentation $d$. }
\label{fig:learningproblem}
\end{figure}
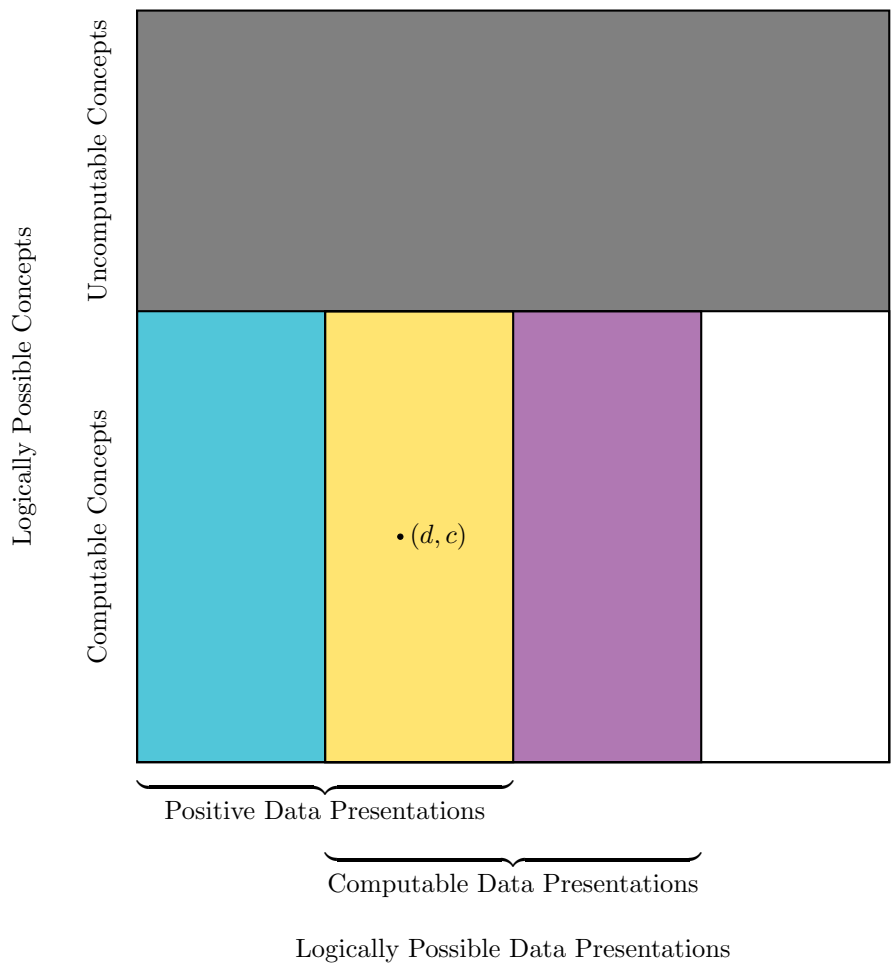

Figure~\ref{fig:learningproblem} visualizes some of the parameters
involved in defining a learning problem.  
In the figure, the $y$-axis represents all possible concepts,
including uncomputable ones. 
The $x$-axis represents all logically possible data presentations:  
computable ones, uncomputable ones, ones with only positive examples, 
ones with only negative examples, 
and ones with both positive and negative examples. 
A learning problem is defined in part by
selecting some subset C of the logically possible concepts
and identifying, for each concept $c$ in C,
the data presentations D$_c$ that a learner is expected to succeed on. 
Then a learning algorithm A can be said to learn the concepts C
from data presentations of kind D if and only if
for all $c$ belonging to C, and
for all $d$ belonging to D$_c$,
it is the case that A($d$)$\approx c$.
Learning problems may also bound the computational resources
that A is allowed to use, possibly as a function of $d$ and $c$.
Readers are referred to \citep{heinz2016} for a survey of different formal learning paradigms. 

Piantadosi cites \citet{ChaterVitanyi2007}
as an example of work that apparently defies these computational laws
because these authors present a paradigm and algorithm
that learns any computable language. 
However, there are two important qualifications to this work that
Piantadosi fails to mention.
First, Chater and Vit\'{a}nyi themselves note that
their result only holds because they
\textit{reduce the instance space of data presentations}.
In particular, the learner of \citet{ChaterVitanyi2007}
is only required to succeed on positive data presentations
``generated by some monotone computable probability distribution'' (p.~138).
This result is actually similar to one obtained by \citet[p.~469]{gold67}
who showed that the class of computable languages is learnable
when the positive data presentations are limited
to ones generated by a particular kind of computable function (Theorem I.7).
Chater and Vit\'{a}nyi's result, like Gold's,
follows in part because the instance space of the learning problem
has been reduced to cases where the data presentations include positive examples
generated by computable processes (see Figure~\ref{fig:learningproblem}).
In other words, Chater and Vit\'{a}nyi's results exemplify
the trade-off mentioned above: 
if one reduces the instance space of the learning problem,
by limiting which \textit{data presentations} learners have to succeed on,
one can expand other aspects of the instance space,
such as the \textit{family of concepts} to be learned.
This observation is not new; it is originally due to \citet{gold67}.

While Chater and Vit\'{a}nyi's reduction of the instance space is significant, however, it is not enough to yield \textit{feasible} learning of all computable languages.
Another foundation of \citeauthor{ChaterVitanyi2007}'s \citeyear{ChaterVitanyi2007}
theoretical result is that they allow
their algorithm to make ``uncomputable'' (p.~136) calculations. 
Incidentally, this is in contrast to Gold's Theorem I.7,
which only considers computable algorithms.
While Piantadosi is optimistic
about what Chater and Vit\'{a}nyi's ``ideal'' learner means in practice,
Chater and Vit\'{a}nyi are more circumspect. 
Setting aside the fact that that
the heuristics needed to bypass the uncomputable calculations
render their theoretical result inert,
they acknowledge that ``real language learning must occur reliably
using limited amounts of data'' and therefore
``a crucial set of open questions concerns
how rapidly learners can converge well enough
on the structure of the linguistic environment
to succeed reasonably well in prediction,
grammaticality judgments and language production'' \citep[155]{ChaterVitanyi2007}.  
In this regard, it is worth mentioning that, to our knowledge,
every learnability result that presents an algorithm which 
``learns'' the class of all computable languages or functions
requires \emph{infeasible} amounts of time and data.
Feasible learning -- that is, learning with limited time and data --
requires navigating a \textit{restricted hypothesis space}.

Even the empiricist-minded \citet[chapter~7]{Clark-Lappin-2011} recognize
that constraining the hypothesis space is likely the best way
to obtain feasible learning results.
They suggest one approach is to ``construct algorithms for subsets
of existing representation classes,
such as context-free grammars'' \citep[149]{Clark-Lappin-2011}. 
In other words, they advocate restricting the class of grammars
targeted by learning algorithms to \emph{subclasses}
which are feasibly learnable.
Such a reduction of the hypothesis space is, at its core,
an innate mechanism akin to those advocated for by proponents
of the Poverty of the Stimulus argument \citep[e.g.][]{nowaketal2002}.

At the end of the day, the results from CLT
clearly and firmly support the Poverty of the Stimulus argument. 
\textit{Even if} LLMs were achieving some sort of ``unconstrained" learning as Piantadosi argues, CLT tells us that this would only be possible \textit{because} they were trained on inhumanly large training data, based on the tradeoffs outlined above. 
Learning from small, feasible amounts of data and computational resources \textit{requires}
constraining the hypothesis space, meaning that \textit{even if LMs eventually succeed} at learning from small data,
it will be because they encode ``non-trivial structural priors
facilitating language acquisition and processing''
as \citet[7]{baroni2022proper} suggests. 
At some level, Piantadosi must understand this,
because he himself suggests (pg. 14) that we might improve training of LMs
on small datasets by
``build[ing] in certain other kinds of architectural biases and principles'' 
or ``consider[ing] learning models that have some of the
cognitive limitations of human learners.''
But what are these biases, principles, and limitations
if not some form of the Universal Grammar
that \citet[19]{piantadosi2023modern} claims LLMs prove ``to be wrong''? 


\subsection{Small Language Models are Anything but Inevitable}

\citet[14]{piantadosi2023modern}  is optimistic that
``our methods for training [LLMs] on very small datasets
will inevitably improve.'' 
This optimism, however, is not well-motivated. 
Firstly, both model size and training size
have seen exponential increases in recent years,
and model performance has increased proportionally \citep{kaplan2020scaling}.
Creating smaller models trained on plausible data
has never been a central goal of natural language processing (NLP) \--- the
field from which LLMs emerge \--- because this field seeks primarily to optimize performance for engineering tasks in a world of increasingly available training data and computing power. 
Secondly, work claiming to successfully train LLMs to achieve human-like performance on human-sized inputs suffers from persistent flaws. Their evaluation methods are weak, often not accurately testing the linguistic phenomena that they purport to or adequately controlling for the presence of side-channel information that might be exploited. Such test sets are particularly susceptible to ``shortcutting'' by neural models, which have a notorious propensity for exploiting unintentional statistical side-channel information across machine learning domains
\citep{narla2018automated,chao-etal-2018-negative,sun-etal-2019-mitigating,
hassani2021societal,wang-etal-2022-identifying}. Hence, it is reasonable to be skeptical that
models' apparent success on existing evaluation metrics reflects
an encoding of the grammatical principle these metrics supposedly test.
We elaborate on these rationales below.


\begin{figure}[t]
    \centering
    \includegraphics[width=0.9\linewidth]{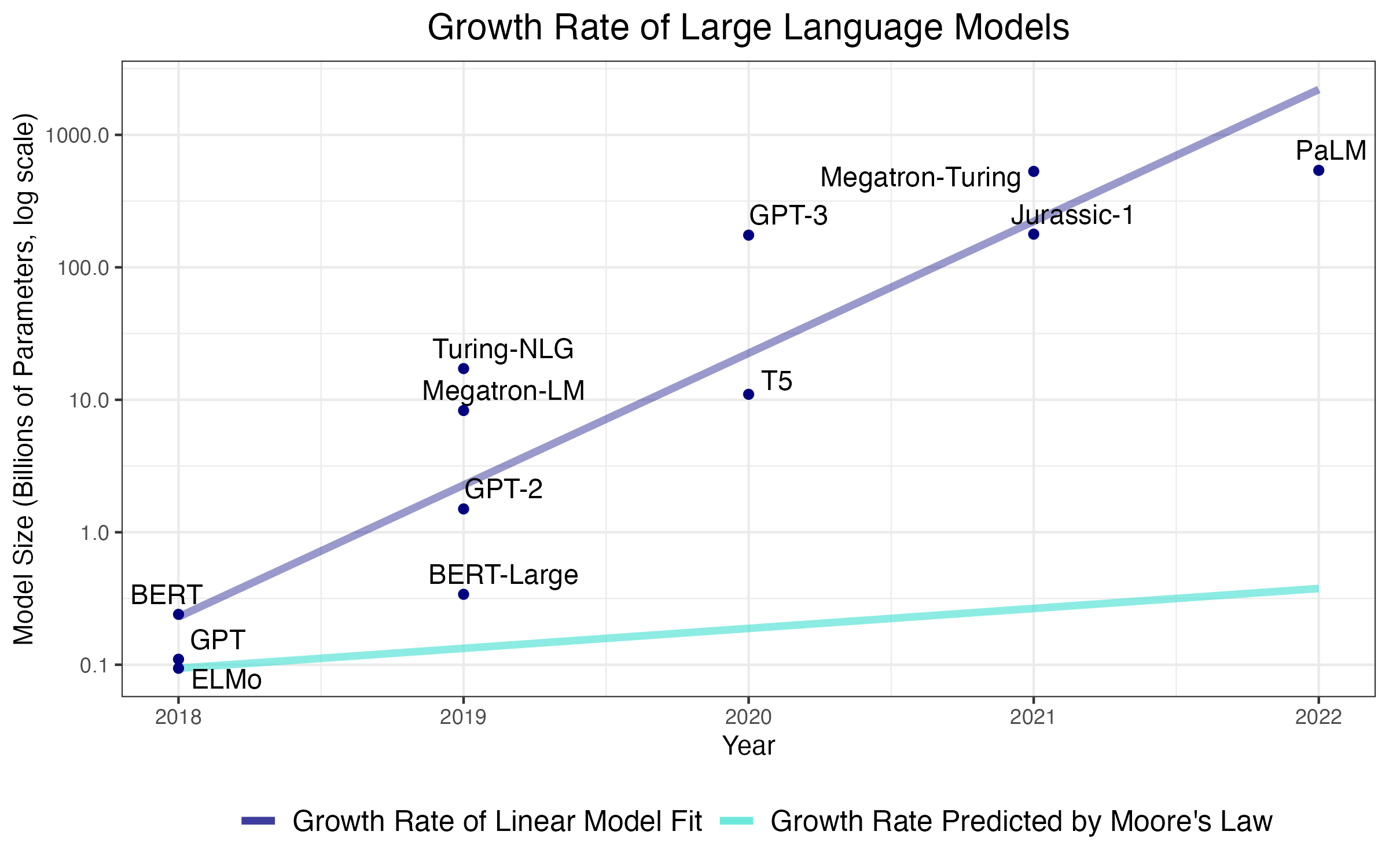}
    \caption{Growth in size of large language models compared to the predicted Moore's Law growth rate, beginning with ELMo.}
    \label{fig:model-growth-moore}
\end{figure}

Over the last four decades, natural language processing
has consistently progressed by consuming more and more data. Just considering the recent history of transformer LMs,
while BERT \citep{devlin-etal-2019-bert} was trained on 3.3 billion tokens,
GPT-3 \citep{gpt3} was trained on 300 billion,
two orders of magnitude increase in only a year, and  thousands of times the input available to a human child. 
This growth in training data has been facilitated by
improved computing hardware and a steady increase in model size:
in the last few years alone, the definition of LLM has shifted
from models with 94 million parameters \citep[ELMo;][]{peters-etal-2018-dissecting}
to 340 million parameters \citep[BERT-Large;][]{devlin-etal-2019-bert},
11 billion \citep[T5;][]{raffel2020exploring},
and 175 billion \citep[GPT-3;][]{gpt3}.
While this pattern has often been dubbed the ``Moore's Law of NLP,"
it actually far \textit{exceeds} the exponential rate of growth
predicted for transistors by the original Moore's Law
(Figure \ref{fig:model-growth-moore}; \citealt{moore}).
Though the exact number of parameters or training data size of GPT-4
are not publicly available, these trends suggest that GPT-4
likely has at least two orders of magnitude more parameters than GPT-3,
and is likely trained on as many orders of magnitude more data; this would match the number of input tokens of perhaps \textit{millions} of human learners, approaching the combined lifetime experience of all 2022 LSAT takers.
What's more, as \citet[14]{piantadosi2023modern} himself acknowledges,
``work examining the scaling relationship between performance and data size
show that at least current versions of the models do achieve their spectacular
performance only with very large network sizes
and large amounts of data \citep{kaplan2020scaling}.''
Put simply, Piantadosi's suggestion that the performance of LLMs with smaller data will greatly improve is overly optimistic, because this has \emph{never} been the trajectory of NLP. 
Nothing short of a paradigm shift would be required
to get researchers chasing state-of-the-art performance
to work with less training data.

We have seen this play out in the field in recent years. While there have been many pushes towards smaller data and more efficient training approaches, such efforts have, as of yet, failed to substantially alter the mainstream course of the field. These pushes include
the 2022 Annual Meeting of the Association for Computational Linguistics (ACL) theme track on ``\textit{Language Diversity: from Low-Resource to Endangered Languages},''\footnote{\hlink{https://www.2022.aclweb.org/post/acl-2022-theme-track-language-diversity-from-low-resource-to-endangered-languages}{https://www.2022.aclweb.org/post/acl-2022-theme-track-language-diversity-from-low-resource-to-endangered-languages}} the upcoming BabyLM Challenge shared task,\footnote{
\hlink{https://babylm.github.io/}{https://babylm.github.io/}} the now-completed DARPA ``\textit{Low Resource Languages for Emergent Incidents}'' (LORELEI) project,\footnote{
\hlink{https://www.darpa.mil/program/low-resource-languages-for-emergent-incidents}{https://www.darpa.mil/program/low-resource-languages-for-emergent-incidents}} and the leaked document from Google discussing the success of open source order-10 billion (`small' in the current era) parameter LLaMA-variants.\footnote{
 \hlink{https://www.semianalysis.com/p/google-we-have-no-moat-and-neither}{https://www.semianalysis.com/p/google-we-have-no-moat-and-neither}}
However, for every small data workshop, there are large data workshops,
like the 2021  Workshop on Enormous Language Models,\footnote{\hlink{https://welmworkshop.github.io/}{https://welmworkshop.github.io/}}
whose call argued that ``naïve extrapolation of these trends
suggests that a model with an additional 3-5 orders of magnitude of parameters
would saturate performance on most current [2021] benchmarks.'' And of course, the continued development and uptake of GPT-family and their LLM competitors, the most data-hungry class of NLP models ever built, is undeniable. 
Large, well-funded research teams are not yet investing
in learning from small data to anywhere near the extent
that they are investing in learning from enormous data. We hope this changes too, but we cannot count on it.


Apparently bucking the trend for exponential growth,
a series of recent papers have claimed to show that
LLMs can already perform well with training data that more closely
resembles a human learner's input \citep[e.g.,][]{huebner2021babyberta,zhang2020you,
warstadt2022artificial, hosseini2022artificial}.\footnote{Authors differ dramatically in how much input they consider human-like. \cite{huebner2021babyberta} train on five million words of American English child-directed speech in their smallest experiments, while \cite{warstadt2022artificial} and the BabyLM Challenge, which also focus on American English, consider 100 million words, corresponding roughly to a ten-year-old's input, to be appropriate. However, English learners express inflectional morphology, agreement, and many major syntactic phenomena
within three to four years, with only a third as much input \citep[e.g.,][]{brown1973first, Slobin2022}. Many of these phenomena are evaluated in popular test suites \citep[e.g.,][]{warstadt2020blimp,huebner2021babyberta}.}
The authors support their conclusions by training models
on smaller, often domain-relevant data sets, such as a pre-processed version \citep{aochildes} of the English subset of the CHILDES  collection of child-directed speech corpora \citep{macwhinney2000childes},
and testing their behavior on grammar test suites
of the kind cited by Piantadosi
\citep[e.g.,][]{warstadt2020blimp, gauthier2020syntaxgym, huebner2021babyberta}.
The reasoning behind such test suites is as follows:
models are presented with grammatical-ungrammatical sentence pairs,
designed to test the model's ability to discriminate
between them according to some carefully chosen syntactic
(or in practice, also semantic or lexical) phenomenon.
Such phenomena include, for example, coordinate structure islands,
long-distance subject-verb agreement,
or an appropriate choice of negative polarity items.
In gradient versions of the tasks, the model assigns a probability
to both sentences in the pair,
and if the grammatical sentence is awarded the higher probability,
the model succeeds at the test.
Alternatively, in binary versions of the task
\citep[e.g.,][]{warstadt2019neural},
a classifier is trained on top of the LLM,
and the resulting model's task is to classify sentences
as grammatical or ungrammatical.

If the goal of these test suites is to show that a model is encoding
knowledge of the grammar, however,  further assumptions are
required regarding the interpretation of the model's outputs. 
Firstly, one must assume that the values
output by the model are, in general, a good reflection of human
acceptability judgments.
However, it is not immediately obvious that this is the case: 
\citet{lau2017grammaticality}, for example, report mixed results
when correlating model predictions with human acceptability judgments.
Similarly, though Piantadosi cites
\citet{warstadt2019neural} as an example of an LSTM matching well
with human judgments, he neglects to address the authors' own conclusions
that the model ``perform[s] far below human level
on a wide range of grammatical constructions'' \citep[625]{warstadt2019neural}.
Indeed, the mid-70\% performance Piantadosi
refers to is accuracy aggregated across the entire test set.
When the MCC, a special case of Pearson's $r$ for Boolean variables, is
measured instead, the model achieves only about 0.3,
compared to humans' 0.65-0.8 \citep[630]{warstadt2019neural}.
Here, it appears that good performance in terms of simple accuracy
does not necessarily reflect human-likeness.
Furthermore, when evaluated on controlled test sets
targeting specific grammatical principles,
performance is extremely mixed.
The best-performing model achieves nearly 1.0 MCC on a basic SVO word-order task,
but only 0.15 on the reflexive-antecedent agreement task.
By contrast, \citet[635]{warstadt2019neural} argue that
``most humans could reach perfect accuracy,'' or an MCC of 1, on
the same task.


But even if one were to set these quantitative concerns aside \--- numbers are likely to go up over time \---
there is a second, more fundamental underlying assumption:
that a model will succeed at these tasks  \textit{if and only if}
it somehow encodes something equivalent to the grammar,
or at least the relevant portion of the grammar. 
If there are other possible explanations for
the success of the model on a given test,
then the tests alone can only tell us about a model's predictive abilities,
not any grammar it may encode.
Unfortunately, this assumption is immediately undermined. Creators of these test sets often fail to control for side-channel information that a model could exploit in order to ``succeed" at the task. 
Since neural models are well-known to make use of such side-channel information, skepticism about apparent successes on these tests is warranted. Put simply, the tests do not convincingly show  that
only a model that has achieved a human-like understanding of language can succeed at these tasks.

Consider, for example, the subject-verb agreement test sentences in BLiMP \citep{warstadt2020blimp}, a large minimal pairs grammaticality test set. The subject-verb agreement test pairs
are intended to test for long-distance agreement dependencies, with the implication that a model which succeeds should employ underlying hierarchical rather than linear representations. In our inspection of the test suite, we find that, for a full two-thirds of these sentences, 
the subject and verb are string adjacent (e.g., ``\textit{Most \textbf{legislatures} \textbf{haven't} disliked children.}'') These sentences do not require a model to encode long-distance agreement. For the remaining third, there is an intervening distractor noun (e.g., ``\textit{A \textbf{niece} of most senators \textbf{hasn't} descended most slopes.}''). However, whether or not there is a distractor, it is always the first/leftmost noun that triggers agreement. 
Thus, a model employing ``agree with the leftmost noun'' would achieve perfect accuracy on the the tests even though it does not leverage anything like hierarchical structural knowledge. Indeed, theoretical linguistics provides us with the tools to more thoroughly test models, a point which we will return to in \S \ref{what-look-for}.

Despite numerous shortcomings of this type, BLiMP is a widely used test set, and will form a portion of the test data in the upcoming BabyLM Challenge.
It was also used by \citet{zhang2020you},
who Piantadosi cites as evidence that
LLMs can learn syntax on relatively small data (10-100 million words).
Due to presence of potential shortcuts in this test set, however,
there is reason to be skeptical of Zhang et al.'s conclusions,
and thus of Piantadosi's subsequent optimism for
the prospects of small LMs.
This critique is not meant to disparage the practical utility of modern LLMs or any move towards smaller LMs,
but it does draw into question this kind of evidence and 
the wide-reaching conclusions that some researchers draw from it.\footnote{Similar points have been made regarding conclusions that can be drawn from artificial language learning experiments \citep{RogersHauser2009, RogersPullum2011, JagerRogers-2012}.}

The presence of potential shortcuts such as ``agree with the leftmost noun" is a particular problem for evaluating ANNs in general, since these models are infamous for
deftly exploiting statistical side-channel information. 
This is not limited to NLP alone.
For example, initially promising results from CNNs trained
to detect and classify skin cancer from images
were overturned when it was shown that
the models were actually classifying according
to the presence of rulers or surgical skin ink markings
\citep{narla2018automated,winkler2019association} in the positive images.
The models found an inadvertent easier correlate
with the task objective and focused on that instead of detecting skin cancer,
creating a potentially life-threatening situation.
Closer to our field, neural models have been shown to exploit the \textit{a priori} likelihood of answers in multiple-choice visual question answering (VQA) tasks \citep{chao-etal-2018-negative}. A totally random baseline is expected to achieve 25\% accuracy on a four-option multiple-choice test, yet the models that were tested achieved 52.9\% accuracy when only exposed to the answers with no paired image or question. Since they achieved 65.7\% when the task was run normally, the bulk of their performance has to be attributed to unintended statistical regularities in the distribution of multiple-choice answers rather than an understanding of the VQA task itself. This clear-cut case may also be relevant to GPT-4's performance on the multiple-choice LSAT. 
There are many other examples in NLP as well, including linear shortcuts in probes of linguistic structural knowledge such as patterns like ``agree with the leftmost noun'' and $n$-gram probabilities  \citep{mccoy-etal-2019-right,kodner-gupta-2020-overestimation} and the unintended exploitation of explicit and implicit social stereotypes in training data \citep{sun-etal-2019-mitigating,thompson2021bias},
among others \citep{wang-etal-2022-identifying}.
Social biases induced by biased training data are so omnipresent that mitigation efforts have becomes a subfield unto themselves, for example spawning a series of workshops at NLP venues.\footnote{\hlink{https://aclanthology.org/venues/gebnlp/}{https://aclanthology.org/venues/gebnlp/}}
Given the litany of unexpected shortcuts that LLMs readily discover,
it is a misjudgment to assume that they will not find and take such linguistic shortcuts on the test sets of the kind cited by Piantadosi, 
unless it is robustly demonstrated otherwise. 

Of course, the existence of these shortcuts does not mean that the LLMs subjected to these tests do not encode human-like linguistic knowledge or do not use such knowledge to solve the tests. But, it also means that success on these tests does not tell us that the models do encode linguistic knowledge either.  Additional careful investigation needs to be done past just showing good performance on evaluation sets. For example, by carefully controlling test sets as in \cite{chao-etal-2018-negative}, we can mitigate \--- though not necessarily remove \--- the opportunity for shortcuts, and by employing probes of the internal state of the models, we gain some understanding of the representations that they employ \citep{belinkov2019analysis, tenney2019you, futrell-etal-2019-neural, liu2019linguistic,  manning2020emergent, linzen2021syntactic, rogers2021primer, pavlick2022semantic, wilcox2022using}. Nonetheless, neither approach is a silver bullet,  as we discuss in \S \ref{not-theory}.

\subsection{Section summary}

Piantadosi bases his attack on nativist approaches to language science on the arguments that LLMs represent ``relatively unconstrained'' learners, and that successfully training such unconstrained learners on small data is not only possible, but inevitable. 
To the contrary, LLMs are in fact \textit{not unconstrained}, 
and  unconstrained learning is not possible from plausible human-sized data with feasible computational resources. 
These conclusions follow from fundamental computational laws 
that are no more violable than the conservation of energy laws in physics. 
Put simply, there is as much chance for feasible unconstrained learning in artificial intelligence as there is for a a perpetual motion machine in physics.
Further, the prospect that engineers building LLMs
will fully embrace small data is unlikely, and even if they do, 
current methods for evaluating such models provide little confidence
that they actually encode the grammar in question, 
rather than exploiting statistical shortcuts to ``pass'' the tests.


\section{Simulation is not Duplication}

What can artificial intelligence tell us about natural intelligence?
On the one hand, models offer existence proofs about
procedures that may underlie some cognitive function.
On the other hand, there is \citeauthor{searle1980}'s (\citeyear[422]{searle1980})
critique of \citeauthor{Turing1950}'s (\citeyear{Turing1950}) imitation game:
``Such intentionality as computers appear to have
is solely in the minds of those who program them
and those who use them, those who send in the input
and those who interpret the output.''
The human tendency to anthropomorphize may cloud our scientific
judgments when it comes to inferring underlying mechanisms of complex systems.
But even if one rejects Searle's position, what does imitation mean? 
As \citet[318]{Chomsky2004Turing} writes, 
``Imitation of some range of
phenomena may contribute to this end [providing insight], 
or may be beside the point, as in any other domain.''

\subsection{Multiple Realizability}

The mere fact that distinct systems exhibit 
the same behavior does not mean that they employ the same internal mechanisms.
\citet{guest2023logical}, who apply this reasoning specifically to the question of ANNs as models of cognition, present an example of two clocks,
which appear identical on the outside, but are different on the inside:
clock A is digital, but clock B is analog.
If only the internal mechanism of clock A is known,
would it be a mistake to conclude that clock B uses the same mechanism,
and thus is also a digital clock?
This is an incorrect conclusion despite their identical appearance and behavior.
Similarly, both planes and birds can propel themselves through the air.
Should we conclude that birds are powered by jet fuel because we know how to build jets but not birds?

Analogies of this sort abound. In a public discussion that Piantadosi participated in following the publicizing of his paper, Rosa Cao of Stanford asked Piantadosi  about
whether identical performance indicates an identical mechanism with
this example:\footnote{\hlink{https://columbiauniversity.zoom.us/rec/share/P_izKvUvrdJkTdkwCnzhGDgs1QEprGlMUjA83KkMjCAfNYY4PROifSW9dAotKI6V.NO4JQuj7anb3HZXj?startTime=1679324349000}{Link to a recording of the public discussion}}
two students pass an exam, but student A cheated,
whereas student B genuinely understood the material. 
Clearly, identical performance on a test does not mean
the same processes have been invoked. 


Each of these examples can be described as a case of \textit{multiple realizability}. That is, similar outcomes can be achieved by superficially similar but underlyingly distinct mechanisms of operation, whether that is the flight of birds and jets, or the grades of honest students and cheaters.
Multiple realizability is a particular problem in cases where the systems under investigation are black boxes. A clock that we cannot open to inspect is a black box, as is the instructor's view on the test preparation strategies of a student. Human cognition and the internals of massive LLMs are largely black boxes as well.
A common strategy for detecting multiple realizability is to shed light on the nature of the black box with other information about its internal mechanisms.
For example, if we are trying to understand
how a bird propels itself through the air, we do not turn to the flight of jets, because we know from other observations that animals do not burn jet fuel. Similarly, we can identify a probable cheater if we catch them passing notes with their neighbor.

\cite{guest2023logical} formalize this reasoning for application to cognitive claims drawn from ANNs. It is an inappropriate application of modus ponens to conclude that a neural model is a cognitive model because it predicts human behavior. Rather, if the neural model is (through additional evidence) a plausible cognitive model, then we should expect it to behave in human-like ways. Thus, we should ask whether what we know about LLMs suggests any underlying commonality to humans.
However, like birds and airplanes, our knowledge of the two systems
suggests that the underlying mechanisms are quite distinct.

We provide  three examples. First, as previously discussed, ANNs rely on many orders of magnitude more data than humans do to achieve their levels of performance. Thus, from the perspective of CLT, a human language learner or LSAT test taker are solving a different learning problem than GPT training for the same tasks.
Second, the learning mechanisms employed by ANNs rely heavily on backpropogation, which neuroscientists believe is a biologically implausible way to pass and update information \citep{lillicrap2020backpropagation,yang2020artificial}.\footnote{Especially at an implementational level, even when an ANN appears to employ a similar problem-solving strategy \citep{zipser1988back,stork1989backpropagation}. A body of literature on biologically implementable equivalents to backpropagation in ANNs exists in both the machine learning and neuroscience \citep[e.g.,][]{mazzoni1991more,balduzzi2015kickback,ahmad2020gait}, but this is primarily focused on computational equivalents or alternatives rather than supporting the notion of standard backpropagation through gradient descent ``backward in time'' as a biologically plausible process. It further emphasizes the implausibility of backpropagation as the term is normally used.}
Third, the success of LLMs depends in part
on the large ``context windows'' and other built-in ``hyperparameters'' that they use.
The context window roughly refers to the length of a sequence of words
these LLMs can access when generating responses to user prompts.
The size of the context windows of state-of-the-art LLMs number in the thousands.
GPT-3, for example, has a context window size of about 2,000,
and people speculate GPT-4 has increased this anywhere from a factor of four to a factor of 20 (nobody knows for sure since OpenAI will not release the details, see \S\ref{not-theory}). There is no sense in which humans have any kind of working-memory counterpart to this, which would require a perfect memory of thousands of recently observed words. 


Continuing the reasoning from \cite{guest2023logical}, one may also apply modus tollens to reason through negative results. If a particular ANN does not behave in a human-like way, then that one is not a good model of cognition. Earlier ANNs consistently under-performed compared to humans, so those particular implementations could be rejected. However, no model is perfect, so the failure of a particular ANN cannot lead us to conclude that ANNs as a class should be rejected too. This unfortunately renders a simple negative  existence proof for a cognitively plausible ANN untenable, just as a positive result cannot be interpreted, in itself, as a positive existence proof. One recurring source of non-human-like behavior in ANNs, however, is their inability to reproduce human-like learning errors even when they achieve high levels of performance.

A classic example of this discrepancy emerged during the Past Tense Debate, a predecessor to the modern debates on the cognitive plausibilities of ANNs which raged in the 1980s and 1990s. The debate was superficially centered on computational models for the acquisition of English past tense inflection, but the fundamental issue at stake was whether connectionist ANNs, with their supposedly \textit{tabula rasa} nature and their distributed representations, could unseat models of inflection representation and learning drawn from prevailing linguistic theory \citep{rumelhart1986learning, pinker1988language, pinker2002past, mcclelland2002rules}. What's new is old, in that sense. 
One major observation from the old debates was divergent behavior between human learners and the ANNs of the time in terms of \textit{overregularization} and \textit{over-irregularization}.

Overregularization, the application of a regular/productive/default pattern to a form that should be irregular (e.g., *\textit{feeled} for \textit{felt}) is robustly attested in observational and experimental studies on the acquisition of the English past tense as well as cross-linguistically. It makes up between 5\% and 10\% of productions in child German \citep{clahsen1993inflectional}, English \citep{marcus1992overregularization,xu1995weird,maratsos2000more,yang2002knowledge,maslen2004dense,Mayol2007}, and Spanish verbs \citep{clahsen1992regular,Mayol2007}. Importantly, overregularization occurs \textit{regardless} of the frequency of the productive process \citep{marcus1992overregularization, belth2021greedy}. Furthermore, overregularization often leads to a \textit{developmental regression} or \textit{U-shaped learning trajectory}: a dip in overall production accuracy when a child learns and begins to over-apply a productive process \citep[e.g.][]{marcus1992overregularization, ravid1999learning, clahsen2002development}. 
Over-irregularization, on the other hand, (e.g., \textit{wing}-*\textit{wang} by analogy to \textit{sing}-\textit{sang}) is far rarer in the same studies, under 1\% in German participles, 0.2\% in English, and 0.01\% in Spanish.

However, the past forty years of debate have shown a persistent failure of ANNs to replicate these human learner patterns. In the first match of the debates, \cite{pinker1988language} already observed frequency-dependent over-generalization in the ANN of \cite{rumelhart1986learning} and showed that the latter's apparent developmental regression was only achieved due to severely unnatural training data presentation. While the raw accuracy of much more powerful modern ANNs dwarfs that of \cite{rumelhart1986learning}, it has been repeatedly demonstrated that these various architectures still fail to produce human-like error patterns. They are still overly frequency-dependent, fail to achieve developmental regressions where appropriate, and do not yield the expected asymmetry in overregularization and over-irregularization \citep[e.g.][]{corkery2019we, mccurdy-etal-2020-inflecting, mccurdy2020conditioning, kodner-khalifa-2022-sigmorphon, kodner2023re}. It is certainly possible that an ANN architecture not yet invented will address these issues, but the persistence of these problems despite dramatic engineering advances in ANN architecture and training suggests the that issue is a more fundamental characteristic of ANNs as a class.

\subsection{Failures are More Informative than Successes}

Given the basic problem of multiple realizability in cognitive science,
it is strange that Piantadosi endorses \citeauthor{warstadt2022artificial}'s
(\citeyear{warstadt2022artificial}) contention that an LLM's failures
are scientifically less interesting than its successes.
Warstadt and Bowman's reasoning is that successes
count as an existence proof that at least some member
of the class of artificial neural networks can solve the task,
while a failure is ambiguously attributable to either
a fundamental weakness of ANNs as a class or the incidentally
imperfect state of the current technology.
This conclusion is wrong for two reasons.
The first is practical: it requires us to accept that the task that
the researcher has adopted to test some property of the ANN
is itself able to discriminate between a success and a failure.
As we have discussed at length, however (\S \ref{small-data}),
current approaches are not convincing.
Warstadt and Bowman's own grammaticality evaluation test suite,
BLiMP \citep{warstadt2020blimp}, for example,
contains many weaknesses (\S\ref{small-data}, \ref{sec:21st}).
Since neural models of all shapes and sizes will exploit
unintended shortcuts in their input in order to take the path of least resistance \citep[\S \ref{small-data}, e.g.,][]{narla2018automated,
  winkler2019association,chao-etal-2018-negative,mccoy-etal-2019-right,wang-etal-2022-identifying}
it is reasonable to conclude that LLMs are likely ``cheating'' on these tests until proven otherwise.
In such cases, their successes do not constitute an existence proof, but rather a pointer to areas which require further investigation.

The second reason is again the logical problem of multiple realizability. Piantadosi
as well as \cite{warstadt2022artificial} draw exactly the wrong conclusion here.
A positive result lends support to an approach that we have external reasons to believe is a plausible model of cognition. It cannot itself be the justification for that assertion. A negative result, on the other hand, is proof positive that this particular model, in this particular learning setting, is not an appropriate model of cognition. Of course, a related model on a related learning problem may be an appropriate model. It would be ideal if positive results could serve as existence proofs and negative results could eliminate whole classes of models, but the universe need not orient itself for our scientific convenience.

\subsection{Section Summary}

\citet{Somers2013} quotes Douglas Hofstadter,
``Why conquer a task if there’s no insight to be had from the victory? Okay, Deep Blue plays very good chess — so what? Does that tell you something about how we play chess? No.'' 
Claiming that the human language faculty is somehow LLM-like because GPT-4 outperforms most test takers on the LSAT is akin to concluding that Kasparov interprets a chessboard like Deep Blue because it beat him, or that birds burn jet fuel because jets out-speed and out-distance small animals. 
But worse, it would be akin to concluding that Clock B is digital \textit{without understanding how digital clocks work}.
What would we gain from such a conclusion?
Certainly neither explanation nor elucidation:
we would simply replace a mystery with a black box.
If the goal of a scientific theory is to provide an explanation,
then it is unclear how a scientific theory of language based on LLMs might provide this,
a point to which we  turn in \S \ref{not-theory}.


\section{LLMs are not a Scientific Theory}\label{not-theory}


Piantadosi argues that ChatGPT's impressive capabilities
mean that it constitutes a theory of language.
He is not the only writer to advocate for adopting neural networks as theories.
\citet{potts2019case}, for example, advocates for adopting deep learning as a theory of semantics.
However, it is unclear what such a theory tells us about language.
Echoing the connectionists during the Past Tense Debate, Piantadosi mentions gradient representations,
the use of word prediction as a learning signal,
and the lack of built-in constraints as elements of such a ``theory.''
However, these are rather general properties, and no specifics are offered.
It is easy to see why: the role of a scientific theory
is to \textit{elucidate and explain} \citep{popper1959logic},
and LLMs largely fail to do either.

Our argument comes in two parts. First, as corporate models,
LLMs violate the best practices of open science and software development,
and their results are neither replicable nor reproducible.
Second, even if the models \textit{were} open-source,
we are far from understanding how they work,
so they cannot presently provide a scientific explanation.
All ANNs do is predict, and prediction is not explanation.
Theories that made relatively accurate predictions historically, like the well-worn example of Ptolemaic epicycles on epicycles,
have turned out to be incorrect, so prediction is not the end-all for judging the success of a theory.
While ANNs certainly have many useful applications \--- fitting
hypotheses to data or carrying out downstream engineering tasks \--- they
cannot constitute scientific theories for the simple reason
that they currently explain very little about language.

\subsection{Corporate ``Science''}

The LLMs of today are a corporate product, not a scientific one.
Industry dominates the creation of LLMs
due to the high financial and compute costs associated with their training \citep{ahmed2023growing}, and the corporations releasing these LLMs are often cagey about the details of their implementation \citep{liesenfeld2023opening}.
We still do not know, for example, the architecture or training data that GPT-4 uses or how often it is updated, or what  kind of hand-tuning or output filters it has, because information about the product is released through public press releases rather than peer-reviewed publications as is standard in linguistics, NLP, and cognitive science.
This lack of clarity means that modern LLMs are neither replicable nor reproducible,
nor can the most recent LLMs be subject to the probes of internal state that earlier LLMs could be.
Of course, this is likely by design: it is a savvy business strategy not to disclose the details of your model,
lest a competitor beat you at your own game.\footnote{It is particularly surprising that Piantadosi does not express concern about this lack of openness and replicability given his past support for strong replicability requirements for scientific research \citep{rieth2013put}.}

Indeed, this illustrates one of the dangers of using corporate models for science:
the goals of the corporations creating the models (i.e., to increase profits) are not the same as the goals of the scientists
trying to probe the models (i.e., to come to a scientific understanding of language).
LLMs are also constantly changing as new edge cases are found and reported,
and many layers of employees are actively engaged in curating training content and guiding the outputs of the models \citep{Time,WSJ}.\footnote{Articles in \hlink{https://time.com/6247678/openai-chatgpt-kenya-workers/}{Time Magazine} January 18, 2023 and \hlink{https://www.wsj.com/articles/chatgpt-openai-content-abusive-sexually-explicit-harassment-kenya-workers-on-human-workers-cf191483}{The Wall Street Journal} July 24, 2023.} 
 Again, these approaches work well for the corporations who own the models,
since they want the LLMs to behave well so they can maximally profit from them.
But these approaches run against the goals of science,
further obfuscating the implementation details of already opaque models.
The different goals of the corporations owning the models and
the researchers probing them should not be taken lightly.

Consider an analogy to buying a used car:
the salesperson wants to make a sale,
and you want to receive information on the details and value of the car.
If you wouldn't trust the salesperson to give you
completely honest information about the details of the car,
why would you trust corporations to do so for their LLMs?
Both have the same ulterior motives:
to make their product look good in order to gain maximum profit.
As anyone who has bought a used car knows, these motives often lead to stretches of the truth.
But there is no Carfax for LLMs: the normal process of peer review has been bypassed by industry press releases and preprint publications. 

Open source LLMs, such as the newly released LLaMa 2\footnote{Meta press release: \hlink{https://about.fb.com/news/2023/07/llama-2/}{https://about.fb.com/news/2023/07/llama-2/}} mitigate some, but not all, of these challenges but as of yet only constitute a fraction of LLM use. Additionally, LLMs billed as open source, including LLaMa 2, are also not nearly as open as their marketing would lead one to believe, suffering from poor documentation, limited or no access to training data or hyperparameters or output filtering steps, and so on
 \citep{liesenfeld2023opening}.\footnote{See \hlink{https://opening-up-chatgpt.github.io/}{https://opening-up-chatgpt.github.io/} for a growing list of open source LLM scorecards supplementing  \cite{liesenfeld2023opening}.}
Both the new corporate mode of publication and the importance of opensourcing models were focuses of the panel ``The Future of Computational Linguistics in the LLM Age'' at the recent 2023 Annual Meeting of the Association of Computational Lingusitics (ACL), one of the largest gatherings of NLP researchers. This transition away from clear, open, replicable, research is clearly a concern for researchers across NLP as much as it as among linguists and cognitive scientists.

\subsection{Prediction Alone is not Explanation}

Even if LLMs were truly open source and all pieces were known,
they would still not function as a theory of language
because it is not well-understood how they work.
\citet[8]{piantadosi2023modern} himself acknowledges
that ``it [can] be hard to determine what's going on,
\textit{even though the theory is certainly in there}.''
But what does it mean for a theory to be hidden in a black box?

A cottage industry has popped up over the last few years within NLP
seeking to understand the inner workings and behaviors of popular ANN architectures.
It goes by many names, including ``explainability and interpretability''
or ``BERTology'' \citep{rogers2021primer},
though the latter term is beginning to show its age
\citep[e.g.,][]{belinkov2019analysis,  tenney2019you, liu2019linguistic,
  manning2020emergent, linzen2021syntactic, pavlick2022semantic}.
This is certainly a step in the right direction, 
but the largest obstacle is that LLMs are, by nature, 
not easy to interpret.
Even when all parameter values are available, which is no longer generally the case for the most powerful LLMs, it is not straightforward to map these to model behavior,
and this problem is only exacerbated as model size increases.
Indeed, Piantadosi himself acknowledges that
``we don't deeply understand \textit{how} the representations these models
create work'' \citep[8]{piantadosi2023modern}.
Understanding what goes on inside these LLMs is a bonafide research problem,
but this hinders, rather than facilitates,
the case for making LLMs a theory of language,
as we explain in \S\ref{sec:21st}. 

While advances in these research methodologies is progressing rapidly,
the field is stuck playing catch-up with ever-evolving and
increasingly opaque and corporate models. There is no equivalent to BERTology for the latest crop of LLMs because we lack the necessary access to probe them in the way we could even a few years ago.
While there is obvious value in such an enterprise for the purposes
of developing an understanding of state-of-the-art research tools,
we ask whether it makes sense to try to explain the human mind 
by studying an ever-cycling menagerie of opaque human artifacts instead. We do not understand how these LLM artifacts work, they are produced by NLP researchers with entirely different goals in mind, and they will be made obsolete as soon as the next big thing is announced.
If the role of scientific theory is to elucidate,
then a theory of language based on LLMs does the opposite.

Even if we could pin down and perfectly probe current LLMs, these prediction machines still fall flat from the perspective of explanation.
To make a classic analogy to the history of science, consider the case of Ptolemy and Copernicus:
while Ptolemy was able to ``fit the data'' \--- in this case,
explain the relative motion of planets \--- within a geocentric perspective,
doing so required complicating his theory.
The introduction of epicycles within epicycles within epicycles,
and the fine-tuning of each planet's epicycle to best match its respective movements, eventually succeeded in fitting the observations of the day exceedingly well,
and indeed \textit{better} than Copernicus's heliocentric theory when it was introduced.
But of course, we now know Copernicus's view of heliocentrism was fundamentally a step in the right direction. Later refinements of the heliocentric model turned out to not only be the correct explanation, but also a far simpler one,
which not only predicts the motion of heavenly bodies but also explains them.
We can see LLMs and other large statistical analyses as behaving like Ptolemy's theory:
they may fit the data well, and they may even provide extremely accurate predictions.
But theoretical linguistics, unlike LLMs, is able to provide a concise,
explanatory account, even if \--- like Copernicus \--- this account cannot predict language use as well.
Again, if the role of science is to provide elucidation and explanation,
then both heliocentrism and theoretical linguistics are scientific theories,
\textit{even if} Ptolemy and LLMs win the prediction game. Piantadosi suggest that a good theory of language ``\emph{is certainly in}'' the LLMs somewhere. But, if medieval astronomers had all taken that perspective on Ptolemy's model, would they have found their way to heliocentrism?

Piantadosi makes his own analogy to physics,
suggesting that an ANN might be used to determine whether gravitational force
falls off with distance or with distance squared.
This analogy, however, confuses a tool for testing
and elaborating a theory with the theory itself.
In his example, the physicist already has two hypotheses
(\citealt[7]{piantadosi2023modern} calls them ``theories"),
both of which are stated with closed-form, easily interpretable,
mathematical solutions ($\frac{1}{r}$ and $\frac{1}{r^2}$, respectively).
In this example, the ANN \--- or indeed,
any other means of making a maximum likelihood estimate \--- is
only used as a tool for fitting a parameterized version of the hypotheses to the data.
Crucially, these hypotheses are generated beforehand
by the underlying theory of Classical Mechanics and are not themselves
a product of the maximum likelihood estimator.
The role of the ANN is simply to fit a parameter to select between
two possible explanations of the data;
it is thus not itself the theory but rather a tool for distinguishing
the predictions of two hypotheses generated by some other theory.

However, Piantadosi does not merely argue that
LLMs are a useful tool for discriminating between hypotheses in theoretical linguistics. He argues that LLMs \textit{themselves} constitute a theory that should \textit{replace} traditional theoretical linguistics. Under such a view, the LLM would not simply be an adjudicator between hypotheses generated by existing theories, as he recognizes,  ``we don't explicitly `build in' the theories under comparison'' \citep[8]{piantadosi2023modern}. However, he fails to recognize that his own analogy, which does build in and test an existing theory's hypotheses, is inappropriate because of this.

If we insist on drawing one further comparison with physics, a more fitting analogy to his argument comes from ANN applications to the three-body problem:
one may observe that, despite generations of effort,
theoretical physics has consistently ``failed'' to produce
practically usable closed-form solutions to cases of this problem (because no such solution exists mathematically).
However, recent approaches predicting the relative motion of
three bodies statistically with ANNs \citep[e.g.,][]{breen2020newton}
have shown great promise.
Under Piantadosi's reasoning,
the apparent superiority of these neural prediction approaches
over traditional theory should render them
a serious alternative theoretical basis for physics.
This is, of course, absurd.
ANNs and other probabilistic approximators are tools for carrying out predictions
when it is too impractical or impossible to deploy a closed-form solution,
not a replacement for the original underlying explanation.
The three-body ANN does not tell us anything about the \textit{theoretical bases}
of the interaction of the three objects.
Similarly, LLMs don't tell us anything about the theoretical bases
of language merely because they make accurate predictions. 
The idea that a theory could be hidden in the approximator somewhere is a category error.

While we argue that Piantadosi has made an error
by calling ANNs a theory of language,
we agree that they have proven successful in serving as predictive models. 
They form the basis of increasingly useful tools for a wide range of practical applications in the sciences and elsewhere.
\citet[9]{piantadosi2023modern} finds the status of LLM research
``somewhat akin to the history of medicine,
where people often worked out what kinds of treatments worked well
(e.g., lemons treat scurvy) without yet understanding the mechanism.''
He also likens the field to ``modeling hurricanes or pandemics'' in which
``the assumptions are adjusted to make the best predictions possible,'' but this is the same category error again. A good predictive model is not the same as a good theory. Models for predicting weather patterns and pandemics are tools in the scientific toolbox.  They are not the theories themselves. 
The theory is our understanding of a mechanism, not merely the body of observations that spur further research.
Theories in meteorology and epidemiology synthesize everything from fluid dynamics
to physiology, along with direct empirical observation
of real-world complex systems, and yes, computational modeling. 

\subsection{Section Summary}

Piantadosi's endorsement of OpenAI's ChatGPT embraces corporate ``science'' and all the practices that it embodies: unaccessible software, data, lack of replicability, and incentives that align with the pursuit of the bottom line and not the pursuit of truth. By emphasizing prediction to the exclusion of understanding, Piantadosi promotes a disappointingly shallow interpretation of science and what it has to offer. In our view, a linguistic theory should provide explanations for linguistic capacities, not merely predict language text. This is largely concordant with the perspective of van Rooij and Baggio on the nature of theory in psychology, emphasizing explanation over prediction, an understanding of capacities over effects, and a theoretical cycle combining verbal and mathematical formalization with empirical study \citep{vanRooij2020theorydevelopment,vanRooij2021theorybefore}. To call LLMs a theory rather than a tool misses all of this entirely.

\section{Why Linguistics Will Thrive in the 21st Century}\label{sec:21st}


In the previous sections, we argued that LLMs cannot 
constitute a scientific theory of language because 
they are largely proprietary and uninterpretable, 
and their focus is on \textit{prediction}, not elucidation or explanation.  
In contrast, however, linguistic theory aims to provide 
an explanatory account of human languages.
By making use of a set of abstract universals, linguistic theory seeks to concisely explain \textit{why} languages are structured the way they are and make testable predictions about grammatical distinctions within and across the world's languages. 
We argue that \textit{even if} LLMs appear to fit the data better than linguistic theory, 
only the latter succeeds as a scientific theory because only it provides an explanation of \textit{why} the relevant patterns arise. 
Indeed, without linguistic theory, there would be no way to test the linguistic capabilities of ANNs. 
Test suites designed for phenomena such as subject-verb agreement or anaphora 
are designed to test whether LLMs encode the \textit{distinctions provided by linguistic theory}. 
Similarly, work probing LLMs often seeks to find evidence for the abstract universals predicted by linguistic theory, for example hierarchical structure. 
Thus, generative linguistics broadly construed is a true scientific theory of language, one which will continue to thrive in the 21st century. 

\subsection{Linguistic Theories offer Explanations}

Consider the simple example of subject-verb agreement, or the difference in grammaticality between ``\textit{I say she walks}" and ``*\textit{I say she walk}" in most varieties of English. Linguistic theories provide interpretable mechanisms for enforcing this formal distinction in terms of \textit{abstract universals} such as hierarchical structure, features, and locality. For example, in Minimalism, a theory of the structural basis of grammaticality, $\varphi$-features of the subject (person, number, gender) are copied to the verb, and syntactic structural locality constraints on the copying mechanism predict the difference in grammaticality of the sentences.
The same theory that distinguishes the sentences above also makes cross-linguistic predictions about the typology of subject-verb agreement. Agreement relies primarily on syntactic structure, not linear order, so we do not expect to find a language in which the verb always agrees with the noun in the third linear position, for example. 
As new evidence regarding the typology of agreement is introduced (e.g., \citealt{van2015uniform}), the theory is updated to account for this evidence; 
new explanations are found and new typological predictions are made and tested. 
This ability to explain and predict 
the syntactic relations of natural language contrasts sharply with the ability of LLMs \citep{moro2023large}. 
To the extent that LLMs show knowledge (i.e., predictive ability) of subject-verb agreement without the possibility of exploiting side-channel information (\S \ref{small-data}), 
they still do not provide a clear explanation as to \textit{why} this difference exists, 
due to their lack of interpretability. 
Consequently, they cannot make the same kinds of cross-linguistic predictions that syntactic theory does.

The distinction between LLMs and linguistic theory outlined above is analogous to Chomsky's argument that Bayesian modeling and similar statistical methods ``won't get the kind of understanding that the sciences have always been aimed at'' but only ``an approximation to what's happening,'' \textit{despite} potentially fitting the data better than theoretical explanations. While we endorse Chomsky's position here, \citet[26]{piantadosi2023modern} quotes him critically, instead arguing in favor of simulating (such as with an LLM) emergent systems as an alternative to a ``Gallilean'' study of capacities that Chomsky, \citet{vanRooij2020theorydevelopment,vanRooij2021theorybefore}, and others endorse.

Piantadosi counters Chomsky's point with the stock market, and example of an emergent system that he argues is ``understood'' through simulation. But, this is another poor analogy. The stock market, and the economy more broadly, are infamously chaotic systems, and financial institutions must continuously pour vast monetary and personnel resources into their efforts to keep predictions up-to-date,  profitable, and secret from the competition.
Nobody should hope for a similar state of affairs in the sciences.
Given that financial modeling makes many people a lot of money, 
it is telling that economic \textit{theories}, not just massive predictive models, still form the basis of economic policy. 
For all the criticisms that can be levied against the United States Federal Reserve, they are still wise enough not to leave us at the mercy of some ANNs.

Piantadosi's second analogy to emergent behavior in beehives is better. However, it is problematic as well, because computational colony modeling does not rely solely on top-down predictive models. Rather, it also incorporates bottom-up explicit mathematical modeling of individual colony members \citep[e.g.,][]{belic1986mathematical,bonabeau1998model,wittlinger2006ant}. While linguistic theories based on emergence and self-organization exist \citep[e.g., exemplar theory:][]{pierrehumbert2003phonetic,ambridge2020against,gradoville2023future}, these resemble the top-down-plus-bottom-up study of insect colonies, not the current state of black box LLMs. Analogous bottom-up studies of individual neurons in LLMs or the impact of individual input tokens on LLMs is hampered by their truly massive size, computing demands, and proprietary nature.

\subsection{Linguistic Theories Tell Us What to Look For}\label{what-look-for}

To even determine whether the linguistic capabilities of LLMs rival those of humans requires explicating what humans' capacities actually are; in other words, it requires a \textit{separate theory of language}. 
Despite their flaws, evaluation suites of the likes of \citet{gauthier2020syntaxgym}, \citet{ warstadt2020blimp}, \citet{huebner2021babyberta}, and others exist \textit{because} we have a linguistic theory that tells us what to look for. The same can be said for evaluation methods that probe representations in ANNs, for example by searching for the presence of hierarchical or long-distance encodings \citep[e.g.,][]{hewitt-manning-2019-structural, tenney-etal-2019-bert,tucker-etal-2021-modified,papadimitriou-etal-2021-deep}.
Consider, for example, the binding principles for anaphors (e.g., \textit{himself, herself, themselves}) introduced in \citet{chomsky1981}, of which Principle A  accounts for the following differences in grammaticality:
\begin{exe}
    \ex[*] {I think she loves myself.} \label{anaph_1} 
    \ex[*] {I love herself.} \label{anaph_2}
    \ex[] {I think she loves herself.} \label{anaph_3} 
\end{exe}
Principle A explains these differences in terms of the same abstract universals as were used for subject-verb agreement: hierarchical structure, features, and locality. 
In its essence, Principle A states that an anaphor must co-refer to another noun in the same sentence (explaining the ungrammaticality of (\ref{anaph_2}), since \textit{herself} has nothing to co-refer with), and that it must co-refer with the \textit{hierarchically closest} eligible coreferent (explaining the ungrammaticality of (\ref{anaph_1}), since \textit{she} is eligible and closer to \textit{myself} than \textit{I} is). Coreference is implicated in Chomsky's and related theories by copying $\varphi$-features (person, number, gender) from the noun to the anaphor, for example by copying \textsc{\{3, singular, fem\}} from \textit{she} to \textit{herself} in (\ref{anaph_3}). Popular grammaticality test suites are designed to test for the encoding of Principle A in LMs,\footnote{Nevertheless, it is not clear that these tests even succeed at evaluating Principle A in the first place (cf. \S\ref{small-data}). BLiMP, for example, contains seven Principle A data sets. GPT-2 achieves 100\% accuracy on the first one, \hlink{https://github.com/alexwarstadt/blimp/blob/master/data/principle_A_c_command.jsonl}{\texttt{principle\_A\_c\_command}}. These, and all BLiMP sentences, were programmatically generated from templates, not extracted from real data, and these templates have introduced unintended regulariteis in the data which could be exploited as shortcuts. We observe at least one such shortcut for solving \texttt{principle\_A\_c\_command}: Every single sentence begins with an optional determiner or quantifier followed immediately by a noun, and its anaphor is always the last word. The exact same ``agree with the leftmost noun'' linear rule that achieves perfect accuracy on BLiMP's subject-verb agreement test sentences would also achieve perfect accuracy here. This data set does not provide a good test of Principle A. 

But least \texttt{principle\_A\_c\_command} requires a model to recognize morphological agreement. The other data sets contain similar or worse faults. \hlink{https://github.com/alexwarstadt/blimp/blob/master/data/principle_A_case_2.jsonl}{\texttt{principle\_A\_case\_2}} can be solved almost perfectly by just checking that the verb immediately following the anaphoric pronoun ends in \textit{-ing}. \textit{Almost} perfectly, because 13 `correct' sentences, all including the verb \textit{skated}, are actually copies of the corresponding ungrammatical sentence, as in ``\texttt{Leslie imagined herself skated around the hospital}.'' 

Other test suites do not fair better. Zorro was designed specifically to test the abilities of LLMs trained on child-like data. However, all of Zorro's \hlink{https://github.com/phueb/Zorro/blob/master/sentences/babyberta/binding-principle_a.txt}{\texttt{binding-principle\_a}} test sentences are similar to but even weaker than BLiMP's \texttt{principle\_A\_case\_2}. The word following the pronoun always ends in \textit{-ing} in the grammatical sentences and never in the ungrammatical. Additionally, the 3rd-to-last word always ends in \textit{-ing} only in the grammatical sentences, and most simply, only the grammatical sentences even contain the substring \textit{ing}. Any of these could contribute to an exploitable shortcut, which raises the question of what models actually do. As such, it is unclear what we can conclude from evaluation on this data, as discussed in \S \ref{small-data}.

}
but without the work of \citet{chomsky1981} \--- and theoretical linguistics more broadly \--- we would not have this principle to test. We might intuitively know that (\ref{anaph_1}) or (\ref{anaph_2}) sound \textit{bad}, and we could identify them as vanishingly rare in language corpora, but we need a theory to explain \textit{why} they are bad. 
Theoretical linguistics gives us this explanation.  
Again, \textit{even if} LLMs can perfectly discriminate sentences like (\ref{anaph_2}) or (\ref{anaph_3}) from sentences like (\ref{anaph_1}), they still do not explain \textit{why} the difference in grammaticality exists. 
Theories make concrete predictions about the \textit{causes} of the difference in grammaticality between sentences, and these predictions can be empirically tested in ways that explicitly control for potential confounds. 

\subsection{Linguistic Theories Make Fundamental Distinctions}

Consider an even more fundamental distinction.
The sentence ``\textit{colorless green ideas sleep furiously}'' was famously introduced by \citet{chomsky1957} to demonstrate the independence of \textit{structural} information \--- the syntax \--- from information about \textit{meaning and interpretation} \--- the semantics. All the bigrams in this sentence \--- \textit{colorless green}, \textit{green ideas}, \textit{ideas sleep} and \textit{sleep furiously} \--- 
are semantically infelicitous (i.e., they make little or no sense. Something cannot be both \textit{green} and \textit{colorless}, \textit{ideas} cannot \textit{sleep}, etc.). Despite this, the sentence is syntactically well-formed and shares an identical structure with plenty of mundane sentences like ``\textit{Fluffy orange cats sleep peacefully}.'' 

Piantadosi is confident that ChatGPT has uncovered this distinction on its own. He touts several sentences prompted from ChatGPT which he believes to be similar to \textit{colorless green ideas sleep furiously} in that they are ``rare but not impossible'' \citet[16]{piantadosi2023modern}. But infelicity is not the same as rarity, and none of Piantadosi's sentences make this crucial distinction. For example, ChatGPT's ``\textit{blue glittery unicorns jump excitedly}'' is not nonsensical in the same way as Chomsky's sentence: there is nothing impossible about being \textit{blue} and \textit{glittery}. If internet art is any indication, \textit{glittery} may be the natural state for a \textit{unicorn}, and it is not at all nonsensical for equine unicorns to \textit{jump excitedly}. Indeed, only one of his examples contains even a possible infelicitous bigram: \textit{clouds dream}. The bigram (with a few interpretations, not all relevant) returns over one hundred thousand hits on Google  (contrast a Google search for ``colorless yellow'' or ``crowded empty plywood'').

Moreover, all of the examples provided by ChatGPT are templatic copies of Chomsky's sentence. Each has the form ``\textsc{Adjective1 Adjective2 Noun Verb Adverb},'' matching the sequence of the original sentence. Across all output sentences (with only slight deviations in the first), the initial adjective constitutes a color term and the second one has to do with being \textit{glittery}, \textit{shiny}, or a related word. It seems, then, that despite producing a canned explanation of Chomsky's sentence that was certainly presented many times over in its training data, ChatGPT has not implemented this explanation in generating new sentences. It does not seem to distinguish \textit{rarity} from \textit{infelicity}, or truly ``understand'' that the distinction made by Chomsky's sentence is the independence of syntactic structure. Rather, like the old ANNs of the Past Tense Debate, it resorts to frequency more than anything.
\citet[16]{piantadosi2023modern} acknowledges that ChatGPT
``does not as readily generate wholly meaningless sentences ... likely because meaningless language is rare in the training data," but that was the point of Chomsky's sentence, which both ChatGPT and Piantadosi seem to have missed.



    

\subsection{Section Summary}

Put simply, without linguistic theory, we do not know what distinctions we expect LLMs to make, nor do we know how we expect them to encode those distinctions. Abstract universals such as hierarchical structure, features, and locality, give computational-level explanations for the patterns observed within and across languages: they explain the differences in grammaticality in sentences used to test LLMs, and they tell us what to look for when probing the internal state of the LLMs. Without linguistic theory, the possibility of testing LLMs is dead in the water. At the same time, however, linguistic theory goes far beyond benchmarking LLMs; it makes testable, interpretable predictions about the computational nature of cognitive linguistic representations and their relationships, explaining the variation that exists in the world's languages.

\section{Conclusion}


Large language models are the current pinnacle of achievement in NLP, and the hype surrounding them is not completely unwarranted. For the first time since the early days of ``automatic language processing'' in the 1950s and 1960s, the outputs of NLP research are of broad, accessible, even transformative, utility for the general population. But that does not mean that every claim regarding their transformative power is warranted as well. Our response to \cite{piantadosi2023modern} questions the role of LLMs in the science of language. Do they constitute a linguistic theory, as Piantadosi argues, or are they a just new and powerful tool? We have argued for the latter position for four main reasons. First, LLMs do nothing to refute the Poverty of the Stimulus argument: they are likely not as unconstrained as Piantadosi and others claim, and \textit{even if they were}, this would only be possible because of the inhumanly massive amounts of training data to which they are exposed. In contrast to LLMs, children are fluent, competent speakers of their native language(s) after relatively little exposure; this is a central mystery of language learning that linguistics as a scientific discipline continues to explore in the 21st century. Second, it is inappropriate to conclude that because an LLM predicts human behavior in some way, it is a cognitive model: simulation is not duplication. Indeed, much of what we know about LLMs \--- and ANNs more broadly \--- suggests that they are in the same kind as relationship to humans as airplanes are to birds. Third, LLMs cannot constitute linguistic theories: they are, at the end of the day, uninterpretable, unaccessible corporate software, and they provide prediction rather than explanation. This point  does not detract from the practical utility of LLMs for NLP, but being a powerful tool does not necessarily make for a powerful theory. Finally, to even determine whether the linguistic and cognitive abilities of LLMs rival those of humans requires explicating what humans' capacities actually are. Ergo, it requires a separate theory of language. We have concluded with a summary of why generative linguistics  provides such a theory: it tells us not only what to look for in our models, but also makes testable, interpretable predictions about the computational nature of linguistic representations and their relationships. 

Our four arguments against LLMs as a theory of language are non-exhaustive. They both complement and elaborate on points made by other authors \citep{katzir2023large,rawski2023modern} and likely more to come. A rejection of Piantadosi's view is not a rejection of progress: the capacity of modern LLMs as NLP tools is still astounding, and their dramatic rate of growth suggests that further progress is on the immediate horizon. But when it comes to why children learn language the way they do, or why certain patterns surface and others do not cross-linguistically, LLMs have little to say. In the 21st century, these questions will continue to be asked and answered by linguistic theory. 

\section*{Acknowledgments}

We thank the Department of Linguistics and the Institute for Advanced Computational Science  at Stony Brook University, which provide a broad, deep, and scientifically rich interdisciplinary environment which synthesizes theory and prediction.  
We are also grateful to Spencer Caplan, Bill Idsardi, Mitch Marcus, Scott Nelson, and Charles Yang for their feedback on drafts of this work as well as Bob Berwick's reading group for valuable discussion which led to this undertaking.
S.P. gratefully acknowledges funding by the Institute for Advanced Computational Science Graduate Research Fellowship and the National Science Foundation Graduate Research
Fellowship Program (NSF GRFP) under NSF Grant No. 2234683. 
Any opinions, findings, and conclusions or recommendations expressed in this material are those of the authors and do not necessarily reflect
the views of the funding agencies or of our colleagues. 

\bibliography{bib}
\bibliographystyle{apalike}

\end{document}